\documentclass[journal]{IEEEtran}
\usepackage{etoolbox}
\makeatletter
\patchcmd{\@makecaption}
  {\scshape}
  {}
  {}
  {}
\makeatother

\usepackage{cite}

\ifCLASSINFOpdf
   \usepackage[pdftex]{graphicx}
   \graphicspath{{../pdf/}{../jpeg/}}
   \DeclareGraphicsExtensions{.pdf,.jpeg,.png}
\else
   \usepackage[dvips]{graphicx}
   \graphicspath{{../eps/}}
   \DeclareGraphicsExtensions{.eps}
\fi

\usepackage{amsmath}
\usepackage{amssymb}
\usepackage{algorithm}
\usepackage{algorithmic}
\usepackage[T1]{fontenc} 
\usepackage{multirow}
\usepackage{siunitx}
\usepackage{color}
\usepackage{parnotes}
\usepackage{booktabs}
\usepackage{caption}

\usepackage{xspace}
\makeatletter
\DeclareRobustCommand\onedot{\futurelet\@let@token\@onedot}
\def\@onedot{\ifx\@let@token.\else.\null\fi\xspace}

\def\wrt{w.r.t\onedot} 

\makeatother
\usepackage[hang,flushmargin]{footmisc}

\newcommand\blfootnote[1]{%
  \begingroup
  \renewcommand\thefootnote{}\footnote{#1}%
  \addtocounter{footnote}{-1}%
  \endgroup
}

\ifCLASSOPTIONcompsoc
  \usepackage[caption=false,font=normalsize,labelfont=sf,textfont=sf]{subfig}
\else
  \usepackage[caption=false,font=footnotesize]{subfig}
\fi

\begin{document}
%
\title{Learning Diverse Fashion Collocation \\ by Neural Graph Filtering}

\author{Xin~Liu ,
        Yongbin~Sun ,
        Ziwei~Liu ,
        and~Dahua~Lin ,~\IEEEmembership{Member,~IEEE,}
        }

\twocolumn[{
\renewcommand\twocolumn[1][]{#1}%
\maketitle
\vspace{-30pt}
\begin{center}
  \centering
  \includegraphics[width=1.0\textwidth]{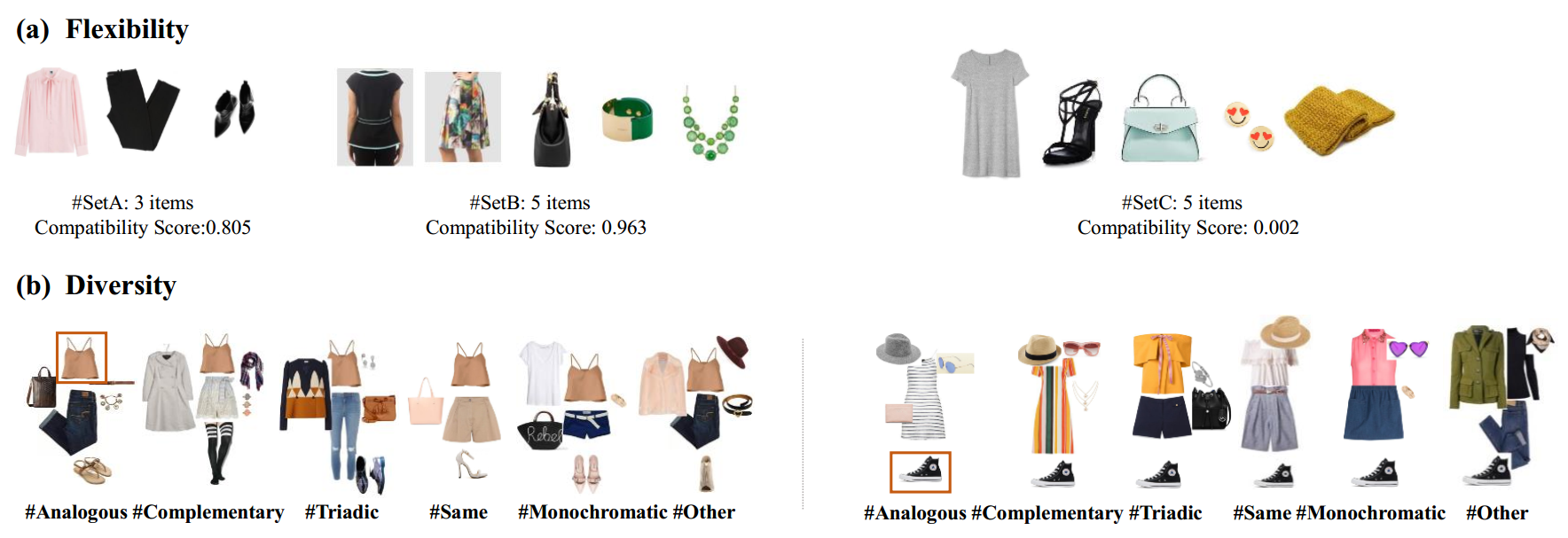}
  \vspace{-10pt}
 \captionof{figure}{\textbf{Flexibility and diversity enabled by our fashion collocation framework.} The first row demonstrates that our proposed neural graph filtering structure is flexible to estimate compatibility scores for various lengths. A higher score indicates higher compatibility (the left two sets), and a lower score indicates a less compatible set (the right set, since usually a tunic dress rarely appears together with a scarf). The second row shows that given one query item (\emph{e.g.}, a top on the left, or a pair of shoes on the right), the recommendation system can generate fashion sets of diverse predefined styles, such as ``Analogous'', ``Complementary'', etc.}
  \label{fig:general}
\end{center}
}]

\blfootnote{\noindent Xin Liu was with the Department
of Electrical and Computer Engineering, Duke University (e-mail: xin.liu4@duke.edu). \\
Yongbin Sun was with Massachusetts Institute of Technology (e-mail: yb\_sun@mit.edu). \\
Ziwei Liu and Dahua Lin are with The Chinese University of Hong Kong.}

\begin{abstract}
Fashion recommendation systems are highly desired by customers to find visually-collocated fashion items, such as clothes, shoes, bags, etc.
While existing methods demonstrate promising results, they remain lacking in flexibility and diversity, e.g. assuming a fixed number of items or favoring safe but boring recommendations.
In this paper, we propose a novel fashion collocation framework, \textbf{Neural Graph Filtering}, that models a flexible set of fashion items via a graph neural network.
Specifically, we consider the visual embeddings of each garment as a node in the graph, and describe the inter-garment relationship as the edge between nodes. 
By applying symmetric operations on the edge vectors, this framework allows varying numbers of inputs/outputs and is invariant to their ordering.
We further include a style classifier augmented with focal loss to enable the collocation of significantly diverse styles, which are inherently imbalanced in the training set.
To facilitate a comprehensive study on diverse fashion collocation, we reorganize Amazon Fashion dataset with carefully designed evaluation protocols.
We evaluate the proposed approach on three popular benchmarks, the Polyvore dataset, the Polyvore-D dataset, and our reorganized Amazon Fashion dataset.
Extensive experimental results show that our approach significantly outperforms the state-of-the-art methods with over 10\% improvements on the standard AUC metric. 
More importantly, 82.5\% of the users prefer our diverse-style recommendations over other alternatives in a real-world perception study.
\end{abstract}


\IEEEpeerreviewmaketitle

\section{Introduction}

\IEEEPARstart{F}{ashion}  plays an important role in human society; people actively engage in fashion trends as means of self-expressing and social connection.
Such enthusiasm drives the rapid growth of fashion industry as well as the demand for intelligent recommendation systems.
In this work, we address the problems of (1) fashion compatibility and diversity analysis, (2) fashion collocation that aims to generate a set of compatible fashion items (\emph{e.g.}, clothes, shoes, bags) conditioning on certain given items.

Automatic fashion collocation is a challenging task, since it has to deal with the inherent complexity among fashion items, and produce favored and creative recommendations that are appealing to users.
With that being said, a successful fashion collocation framework should be featured with two desired properties: \textbf{\emph{flexibility}} (\emph{i.e.}, support fashion item sets with varying lengths and categories, such as \{clothes, shoes, pants, sunglasses\} or  \{clothes, shoes, pants, bag, hat\} ) and \textbf{\emph{diversity}} (\emph{i.e.}, generate diverse fashion collocation styles).
However, prior arts fail to meet these two criteria.

Existing fashion recommendation systems can only accept: \textbf{1)} the fashion set with fixed length, like the four-garment set \{tops, outerwear, bottoms and shoes\} ~\cite{Cucurull2019ContextAware}, or \textbf{2)} limited categories by discarding less popular categories such as accessories, bags, and hats~\cite{hu2015collaborative,li2017mining,YangAAAI2019Trans}. 

Though some deep learning based systems like~\cite{han2017learning}, \cite{liuLQWTcvpr16DeepFashion} and \cite{VasilevaECCV18FasionCompatibility} support flexible inputs, their performance heavily depend on the order of input and lack efficiency. 
~\cite{Hsiao2018Wardrobe} and \cite{lu2019learning} consider fashion versatility and compatibility; however, both methods require users' personal information as training data and cannot offer diverse style recommendation automatically.

To overcome these challenges, we propose a unified framework in this paper, \emph{Neural Graph Filtering}, to achieve compatibility, diversity and flexibility requirements of fashion collocation.  
Working towards this goal, we observe that the correlation between fashion items plays an important role to affect recommended results. 
This makes sense, because the compatibility and diversity naturally come from the coexistence of multiple garments, and are not meaningfully defined for any individual garment.
To better model such \emph{inter-garment} relationships, we choose to use graph structures with the nodes instantiated by garments, and express their interrelationships by edges connecting them.
In our experiments, the node is represented by the visual embedding of a garment image.
Since the \emph{inter-garment} relation is important, the proposed model operates on graph edges, rather than nodes like regular graph networks, to hierarchically aggregate inter-garment information.
Our experiments demonstrate the effectiveness of such edge-centric operations.
Moreover, the graph structure provides us with a flexible way to handle garment sets with various sizes and categories by simply adjusting the nodes and their connected edges in the graph, which fulfills our flexibility requirement naturally. 

However, a graph constructed in this flexible way presents intrinsic irregularity: the number of its constituting nodes varies for different garment sets, and any permutation of their ordering does not change the underlying structure of the graph.
To better impose the flexible representation ability of a garment graph and handle its irregularity, we implement permutation-invariant symmetric functions to aggregate edge feature vectors along the information forward propagation process.
%
Additionally, to demonstrate that our model supports diverse fashion styles, we further split the existing dataset into distinct styles based on color theory, and include an optional style classifier to consider different styles when making recommendations.

During the collocation information propagation process, we also observe that fashion items are distributed in an highly imbalanced manner across styles in the existing dataset. 
For example, the instances for the style \emph{``analogous''} are
about $150$ times more than those for the style \emph{``monochromatic''}.
We show that our graph filtering network can handle such imbalanced distributions naturally, and further adapt the \emph{focal loss}~\cite{lin2017focal} to balance the recognition of the easily recognized styles and hard recognized styles when training the style classifier.

The main contributions of this work lie in three aspects: \textbf{1)} \emph{The concept of flexible and diverse fashion collocation}: Our proposed fashion collocation system has two appealing properties: a) support both inputs/outputs with flexible lengths, and b) generate fashion sets with diverse styles, as shown in Figure \ref{fig:general}. \textbf{2)} \emph{Novel framework of neural graph filtering}: We demonstrate that the graph structure that explores the inter-garment relationship is more suitable for fashion compatibility learning. Our empirical study shows that this approach can diversify the recommendation results and also improve the performance on the tasks ``fashion compatibility prediction'' and ``fill-in-the-blank'' compared to the state-of-art methods. \textbf{3)} \emph{Newly proposed benchmark and evaluation protocols}: We re-organized the prior \emph{``Amazon Fashion Dataset''} as an unseen fashion collocation dataset to suit our new evaluation requirement/protocol, which comprises of different styles for diversity learning and evaluation.
Code and dataset will be released to facilitate future research.

\begin{figure*}[t]
\centering
\includegraphics[width=1.0\linewidth]{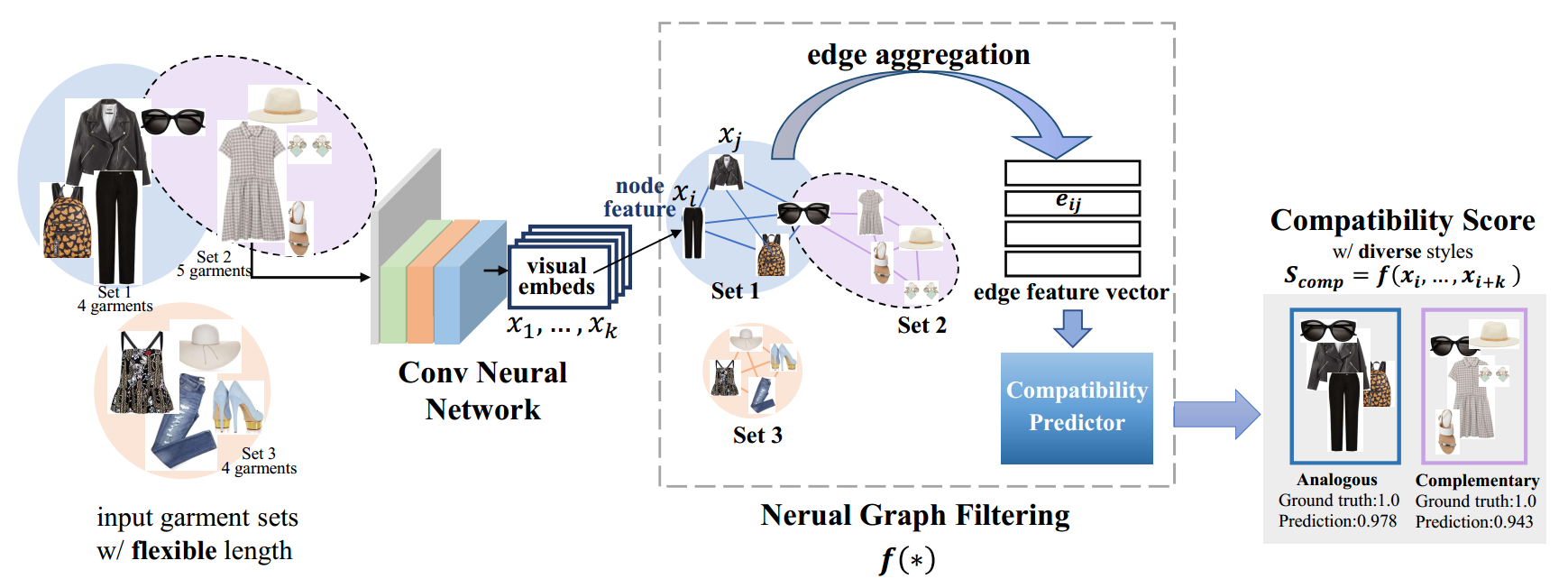}
\caption{\textbf{The Overall Framework of Diverse Fashion Graph Filtering}. We firstly use the convolutional neural networks to extract the visual embeddings of the input garment sets with \textbf{flexible} length, and then consider each visual embedding as a node input to the neural graph network, which not only computes the node features, but also implements edge feature aggregation. Note that one node could appear in several collocations. Afterwards a compatibility predictor calculates the compatibility scores for \textbf{diverse} styled garment sets.}
\label{fig:framework}
\end{figure*}
\section{Related Work}

\subsection{Fashion Collocation}
Prior works on fashion recommendation mainly suffer from constrained settings; for instance, ~\cite{YangAAAI2019Trans} only accepts garments of four categories.
Though Bi-LSTM~\cite{han2017learning} could accept flexible input considering its structure, its performance heavily depends on the order of input, since the sequence-to-sequence-based LSTM model is adopted here.
On the other hand, the paradigm that considers the fashion recommendation as the traditional image retrieval problem that outputs the desired garment one by one purely based on the pairwise relationship, like ~\cite{liuLQWTcvpr16DeepFashion, liu2016fashion} and \cite{VasilevaECCV18FasionCompatibility}, lack efficiency and accuracy in real-world usage.

Not only the lack of flexibility cannot be handled, their recommendation results also ignores diversity.
Existing methods mainly model the compatibility between garments as \textit{visual similarity}, thus almost all their recommended items appear similar, and sometimes boring.
Yet, a diverse fashion collocation system should be able to handle flexible garment set and suggest diverse fashion styles.
Following this track, to go beyond simple visual similarity, \cite{Hsiao2018Wardrobe} represents compatibility by jointly incorporating users' preferences to construct a recommendation system.
\cite{lu2019learning} learns binary code for personalized fashion outfits recommendation. 
However, their work require users' shopping history to measure their preferences that are not easily obtained, and cannot offer specific style suggestions. 
Therefore, to our knowledge, no prior fashion recommendation work has achieved both flexible and diverse requirements. 
Our work is the first to implement a style-respected fashion collocation system and cater to diverse aesthetic preferences without the need of accessing user information. 

\subsection{Graph Neural Networks}
Recently, graph neural networks (GNN) extend standard spatial convolution to graphs.
Compared to spectrum based approaches~\cite{shuman2012emerging,bruna2013spectral,defferrard2016convolutional,kipf2016semi,levie2017cayleynets}, spatial methods directly operate in the feature domain and hierarchically aggregate local information in various manners, including diffusion \cite{atwood2016diffusion} and anisotropic patch-extraction processes \cite{boscaini2016learning}, polar coordinate representations \cite{masci2015geodesic}, learnable local pseudo-coordinate patches \cite{monti2017geometric}, dynamically embedding feature spaces \cite{simonovsky2017dynamic,verma2017dynamic,verma2018feastnet,wang2018dynamic}, and differentiable functional maps \cite{litany2017deep}.
Aside from them, other spatial convolutional strategies have also been proposed, including attention \cite{velivckovic2017graph} and message passing \cite{gilmer2017neural,schutt2017quantum} graph nets. Applications using graphs to capture pairwise relationships can be found in \cite{qi2018learning,shen2018person,vicol2018moviegraphs,wang2018videos}. 

Though ~\cite{Cucurull2019ContextAware} attempted to deploy the graph-like structure for fashion recommendation systems; in the end they created outfits from the database by grouping between 3 and 5 products. Furthermore, this work only used a subset of length 3 of the original outfit for evaluation. Therefore, they have not fully exploited the structural advantage for graph neural filtering to realize flexible fashion collocation.

\section{Our Approach}
We build the proposed fashion collocation system based on graph structures with nodes instantiated by visual embeddings of garment images and edges modeled by the inter-garment relationships. 
The edge information is aggregated and passed along the forward propagation process to calculate a compatibility score for a given garment set, describing how suitable all the items within the set match each other.
When calculating the compatibility score, inter-garment relationships for all different styles are considered.
A highly recommended garment set produces a high compatibility score. 
The overall pipeline is shown in Figure \ref{fig:framework}.
 
\begin{figure}[t]
\begin{center}
   \includegraphics[width=0.8\linewidth]{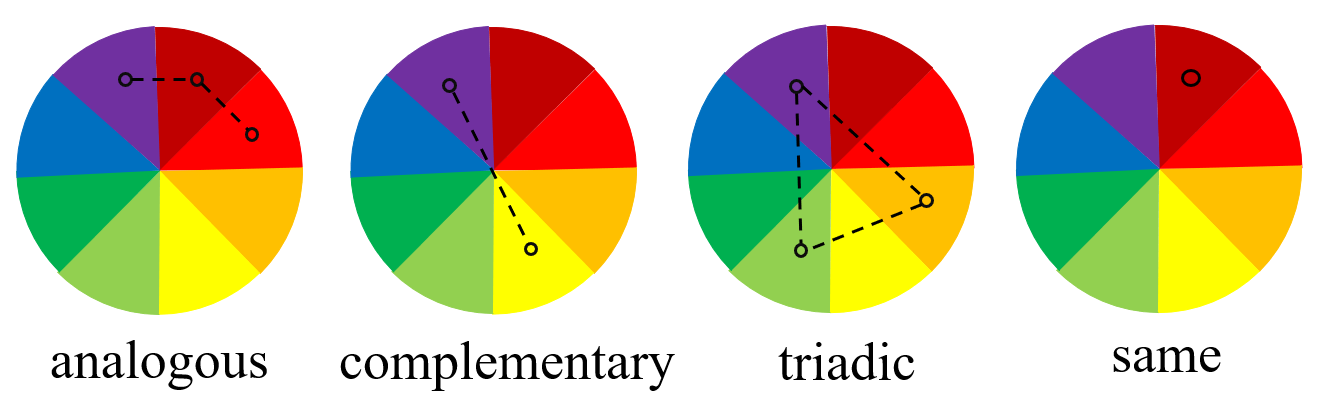}
\end{center}
\vspace{-5pt}
   \caption{ \textbf{Fashion Style Theory}. We denote the garment sets with 6 different styles based on the color theory: ``analogous'', ``complementary'', ``triadic'', ``same'',  ``monochromatic''(various amounts of black and white) and ``other''(no explicit color pattern).}
\label{fig:palette}
\end{figure}

\begin{figure*}[t]
\centering
\includegraphics[width=0.9\linewidth]{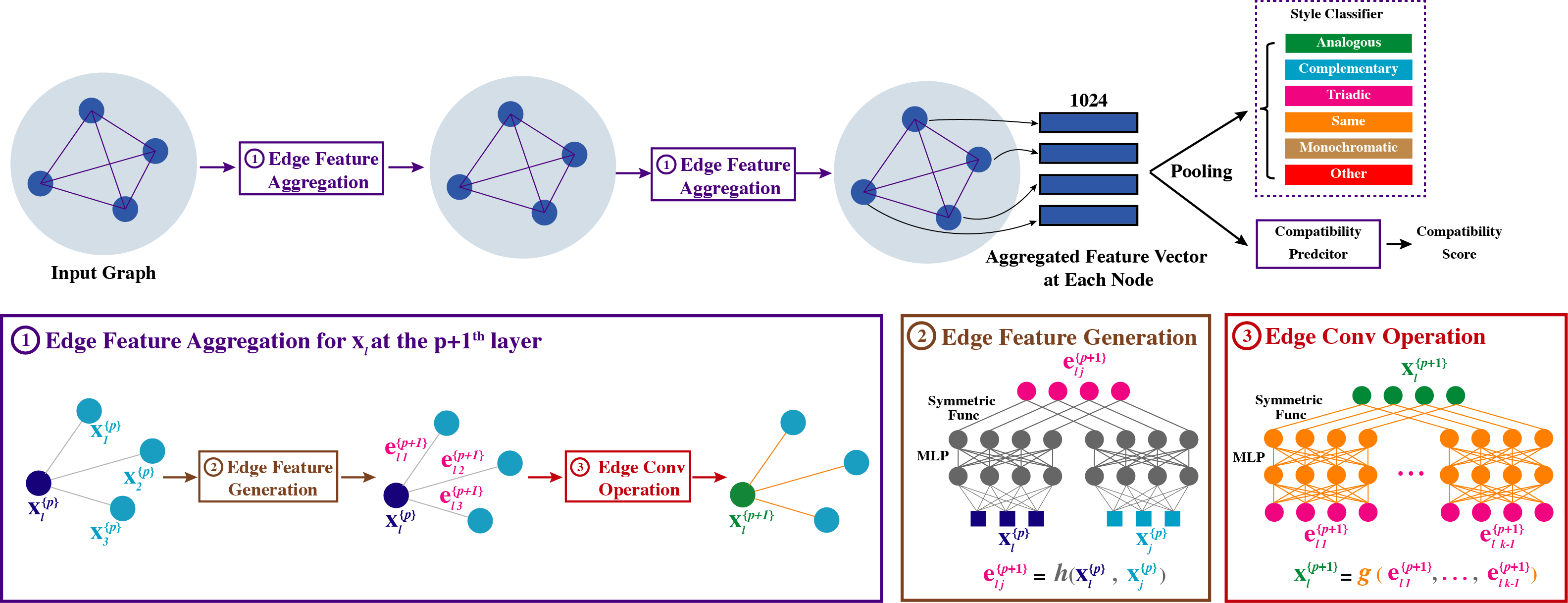}
\caption{\textbf{Architecture of Neural Graph Filtering}. \textbf{Top:} Demonstrate graph edge filtering operation at one layer. This operation aggregates all the edge information connecting to the node under consideration. \textbf{Bottom:} The graph network architecture constructed using hierarchical edge feature aggregation operations. In the last layer, edge information gathered at all the nodes are pooled to compute a compatibility score, and an optional fashion style distribution for a compatible garment set.}
\label{fig:graph_network}
\end{figure*}

\subsection{Style-Diversified Fashion Compatibility}

Fashion compatibility measures whether garments complement one another across visual categories~\cite{han2017learning}. To conquer the tedious fashion tastes, it is necessary to separate the fashion compatibility learning from the directly visual similarity learning, and to evaluate it on diverse-styled angles.
Designers and artists create five major combinations of color palettes according to the color theory~\cite{malacara2011color}, shown as Figure .\ref{fig:palette}, 
that are 
(1)``analogous'': garments with adjacent colors that generate a harmonious feeling;
(2)``complementary'': garments with opposite colors that grab the attention;
(3)``triadic'': garments with colors distributed in an equilateral triangle and thus create a well-balanced feeling;
(4)``same'': multiple garments with exactly same color that generate an integrated feeling;
(5)``monochromatic'': garments with various amounts of white and various amounts of black, which give very modern design in vogue.
We assign all the matched garment sets in the dataset into one of those five distinct fashion styles, and define the remaining ones as ``other". Our model will learn the compatibility relationship from the six styles.

\subsection{Visual Embedding}
We extract the visual embedding of a garment to project it from image space to node space in the graph. Besides, we want the visual embeddings of the items within the compatible sets to gather closer, while the items out of the compatible sets distant from each other.
A favorable approach is to building a triplet relationship for the input anchor item, minimizing the distance between the compatible visual embedding pairs and maximizing the distance between the imcompatible pairs.
Specifically, we compute the visual embedding via a convolutional neural network and denote it as $x_i$, thus a set of garments of size $k$ can be represented as $\{x_i^{u}, x_{i+1}^{v},..., x_{i+k-1}^{r}\}$, where the superscript $u$, $v$ and $r$ represent distinct garment types, such as tops, hats and bottoms.
Among them, any two items $(x_i^{u}, x_{i+l}^{v})$ within the compatible set constitute a positive pair, $1\leq l \leq k-1$, then we randomly sample another item of the type $u$ to constitute a \textit{absolute} negative pair $(x_i^{u}, x_m^{u})$, meaning two items of the same type cannot appear in one set.
We also sample an item of the type $v$ out of the compatible sets to constitute a \textit{relative} negative pair, $(x_i^{u}, x_j^{v})$, meaning two incompatible items of the different types cannot appear in a set.
The distance in these three pairs are given by:

\begin{equation}
\begin{split}
    &d_{\text{pos}} = d(x_i^{u}, x_{i+l}^{v})\\
    &d_{\text{abs\_neg}} = d(x_i^{u}, x_m^{u})\\
    &d_{\text{re\_neg}} = d(x_i^{u}, x_j^{v})
\end{split}
\label{eq:eq1}
\end{equation}

We summarize the distance of the two negative pairs as 
\begin{equation}
 d_{\text{neg}} = \alpha d_{\text{abs\_neg}} + (1-\alpha)d_{\text{re\_neg}}
\label{eq:eq2}
 \end{equation}
 where $\alpha$ means the ratio of absolute negative pairs and relative negative pairs, $0\leq \alpha \leq 1$, and $\alpha$ is 0.5 in our setting.
 
Thus, the triplet relationship is constructed for the anchor item $x_i^{u}$. The triplet loss function is given by:
 \begin{equation}
  loss(i,j,m,l) = \max \{{0, d_{\text{pos}}-d_{\text{neg}} +\mu}\}
  \label{eq:eq3}
 \end{equation}
where $\mu$ is some margin.

\subsection{Graph Edge Filtering}
Given the extracted visual embeddings of garments, we next estimate a compatibility score for a set.
For notation convenience, we ignore the categorical superscript and denote $x_l$ as the embedding of the $l^{th}$ garment, thus a garment set can be expressed as $\{x_1, ..., x_{k}\}$.
The problem from this point on can be formulated as 
\begin{equation}
 s_{comp} = f(x_1, ..., x_k)
\label{eq:comp_model}
 \end{equation}
 where $s_{comp}$ is a scalar compatibility score and $f()$ is a function to be modeled by the proposed graph neural net.

To understand all the pair-wise relationships, we construct a fully-connected graph $\mathcal{G} = ( \mathcal{V}, \mathcal{E} )$, where $\mathcal{V} = \{x_1, ..., x_k\}$ and $\mathcal{E} \subseteq \mathcal{V} \times \mathcal{V}$ are vertices and edges, respectively. 
\subsubsection{Edge Feature Generation}
To better model inter-garment relationships within a graph, we create edge features in a learnable manner: for a node pair, $x_i$ and $x_j$, its edge feature, $e_{ij}$, is obtained via a non-linear function parameterized by some learnable parameters, denoted as $h(x_i, x_j)$.
Since the ordering of an item pair does not change the underlying inter-garment relation, the edge connecting a node pair should be undirected, suggesting that $h()$ should be a symmetric function invariant to the ordering of its arguments, such as pooling operations.

\subsubsection{Edge Conv Operation}
To impose the generated edge features while handling the irregularity of graph, we adapt standard convolution on graph edges in an $1\times1 \, Conv$ manner, followed by another symmetric function to aggregate the resultant feature maps of all edges. 
This type of edge convolution, dubbed \emph{EdgeConv}, summarizes all the edge information associated to a node.

Inspired by the fact that it is important for the information to be diffused across the whole graph such that each garment node can gain enough \textit{within-set} knowledge for determining compatibility, we aggregate all the associated edge features for each node, use them to further construct edge features, and apply \emph{EdgeConv} operation on them again, resulting in stacked \emph{EdgeConv} layers.
Following this iterative way, the information can be fully shared during the forward process.
Without loss of generality, we formulate \emph{EdgeConv} by a symmetric learnable function, $g()$, and express the feature of the $l^{th}$ node in the $p + 1^{th}$ layer processed by \emph{EdgeConv} as:
\begin{equation}
 \begin{split}
	x^{\{p + 1\}}_l  &=  \mathop{g}_{j: (l,j) \in \mathcal{E}}  (..., e^{\{p + 1\}}_{lj}, ...)  \\
			      &=  \mathop{g}_{j: (l,j) \in \mathcal{E}}  (..., h(x_l^{\{p\}}, x_j^{\{p\}}), ...)
 \end{split}
\label{eq:graph_model}
\end{equation}
where $\{x_j: (l,j) \in \mathcal{E}\}$ are node set connected to node $x_l$.

After the last layer, we hierarchically pool the aggregated feature at each node across feature channels, and add a compatibility predictor to compute a compatibility score, ranging between 0 and 1 through a Sigmoid layer.
Optionally, we could add another \emph{style classifier} module to categorize the fashion style for a \emph{compatible} garment set via a Softmax layer using the shared pooled feature. 
The complete process is shown in Figure \ref{fig:graph_network}.

\subsection{Imbalanced Collocation Handling}
We observe imbalanced distributions among different garment items and styles in existing datasets. 
For example, at style level, garment styles favored in common sense usually contain more samples; at item level, garments of common types(e.g. tops) usually appear more frequently than uncommon ones(e.g. hats). 
Fortunately, such imbalance problem can be naturally handled by our graph structure, because the proposed graphical model operates on edges, which encode inter-garment relationships, rather than individual nodes that only contain isolated item information. 
In this way, even though some garment types appear more frequently than others, their inter-garment relationships are dispersed by less-frequent garments associated to them, leading to a more balanced inter-garment distribution which the model actually works on.
Furthermore, working on such inter-garment relationships also allows the proposed model to focus specifically on the internal compatibility within and between garment pairs, as opposed to treating a garment set as a whole, thus the negative effect due to imbalanced garment set style distribution is also reduced.

To better handle the style-level imbalance problem for the optional fashion style classifier, we add focal loss factors \cite{lin2017focal} to each of its cross-entropy losses:
\begin{equation}
\mathcal{L} = \sum_{i=1}^{M} y_i (1 - p_i)^{\gamma}  \log{(p_i)}
\label{eq:loss}
 \end{equation}
 where $M$ is the number of style classes, $p_i$ is the predicted probability for class $i$, $y_i$ is the ground truth style indicator ($0$ or $1$), and $\gamma$ is a hyperparameter ($\gamma = 0.5$ in our settings).

\section{Amazon Fashion Dataset}
Previous fashion recommendations merely conduct experimetns on \textit{Polyvore} ~\cite{han2017learning,VasilevaECCV18FasionCompatibility,YangAAAI2019Trans} since no fashion dataset else contains sets of matched garments, and thus causes serious overfitting and limited transferability.
In such case, we reorganize another dataset, \textit{Amazon Fashion}. It is the subset crawled from \textit{Amazon.com}, firstly proposed in ~\cite{mcauley2015image,KangICDM2017Visual} to recommend visually-related garments, \emph{e.g.}, given a white T-shirt to retrieve a T-shirt with similar color or pattern. 
We hereby reconstruct the dataset into about 60,000 compatible fashion sets, and further put efforts to divide and annotate them into diverse fashion styles.

\noindent\textbf{Diverse Styles Annotation.}
It includes six representation fashion categories (men/women's tops, bottoms and shoes) that appear in five main occasions (casual, sports, work, night and special). While there is no ground truth to guide garments matching in this dataset, we manually construct the sets of garments according to their categories and occasions, \emph{e.g.}, women all-body matches women shoes at ``night'' occasion; men tops, men bottoms and men shoes at ``sports'' occasion match with each other. Furthermore, we split it into six different styles the same as \textit{Polyvore} to increase its applicability and diversity.
Table \ref{table:dataset} shows the detailed distribution of \textit{Amazon Fashion Dataset}.

\begin{table}
 \caption{ \textbf{Dataset Statistics}. We split {\em Polyvore} and {\em Amazon Fashion} into six different styles for fashion diversity learning, both of them show obvious imbalance.}
 \vspace{-2pt}
 \scriptsize
\begin{center}
  \begin{tabular}{c | c c | c c}
    \toprule
      \multicolumn{1}{c}{} &
      \multicolumn{2}{c}{\textbf{Polyvore}} &
      \multicolumn{2}{c}{\textbf{Amazon Fashion}} \\
    \midrule
   {Style} & {train} & {test} & {train} & {test} \\
    \midrule
    Analogous         & 32082 & 6071  & 9104 & 2681 \\
   Complementary & 4018 & 777  & 784 & 232 \\
    Triadic                & 952 & 161 & 238 & 72 \\
    Same                 & 660 & 128  & 7558 & 2107 \\
    Monochromatic  & 208 & 37  & 19732 & 5617 \\
    Other                  & 15386 & 2826 & 5881 & 1661 \\
    \midrule
    Total                   & 53306 & 10000  & 43297 & 12370 \\
    \bottomrule
  \end{tabular}
 \end{center}
 \vspace{-2pt}
  \label{table:dataset} 
\end{table}

\begin{table*}
\scriptsize
\centering
\caption{Comparison with others methods on standard benchmarks: Polyvore (with overlap) and Polyvore-D (no garment overlap). ``AUC'' indicates the AUC of compatibility score, ``FITB'' indicates the fill-in-the-blank accuracy. Our neural graph filtering not only obtains the best performance on both tasks, the human evaluation (``H'' in the table) via a user study also shows our approach is most perceptually appealing.}
\vspace{5pt}
    \begin{tabular}{c | c c | c c |c|c| c c | c c}
    \toprule
    \multicolumn{1}{c}{dataset} &
    \multicolumn{2}{c}{Polyvore} &
    \multicolumn{2}{c}{Polyvore-D} &
    \multicolumn{1}{c}{} &
    \multicolumn{1}{c}{} &
    \multicolumn{2}{c}{Polyvore} &
    \multicolumn{2}{c}{Polyvore-D} \\
    \midrule
    {Metric}  &{AUC} & {FITB}  &{AUC} & {FITB}& {H.(\%)} & {} &{AUC} & {FITB}  &{AUC} & {FITB}  \\
    \midrule
    \midrule
    Bi-LSTM~\cite{han2017learning} & 0.65 & 39.7 & 0.62 & 39.4 & 5.0  & Euclidean Distance & 0.85 & 54.7 & 0.82 & 53.4 \\
    CSN~\cite{Veit2017Conditional} & 0.83 & 54.0 & 0.82 & 52.5 & 0 & Imbalanced Collocation Handling & 0.85 & 55.1 & 0.83 & 54.2\\
    TransNFCM~\cite{YangAAAI2019Trans} & 0.75 & - & - & - & - & Baseline (Node) & 0.92 & 55.3 & 0.84  & 47.8 \\
    Wardrobe~\cite{Hsiao2018Wardrobe} & {\color{blue}0.88} & - & - & - & 7.5 &Baseline (Edge Max Pooling)& {\color{blue}0.93} & 57.7 & {\color{blue}0.87}  & 52.8 \\
    Type Aware~\cite{VasilevaECCV18FasionCompatibility}  & 0.86 & {\color{blue}56.2} & {\color{blue}0.84} & {\color{blue}54.9} & 5.0 & Baseline (Edge Avg Pooling)& {\color{blue}0.93} & {\color{blue}58.0} & 0.86  & {\color{blue}53.8} \\
    \midrule
    \textbf{Neural Graph Filtering (Ours)} & {\color{red}0.94} & {\color{red}58.8} & {\color{red}0.88} & {\color{red}55.1} & \textbf{82.5} & \textbf{Neural Graph Filtering (Ours)} & \textbf{0.94} & \textbf{58.8} & \textbf{0.88} & \textbf{55.1} \\
    \bottomrule
    \end{tabular}
 \vspace{-8pt}
 \label{table: ablation study}
\end{table*}

\section{Experiments}

\subsection{Standard Benchmark Datasets}

\noindent\textbf{Polyvore.} 
We use the standard Polyvore dataset~\cite{han2017learning,VasilevaECCV18FasionCompatibility} for evaluation, which are annotated with outfit and item ID, fine-grained item type, text descriptions and outfit images.
To learn diverse fashion compatibility, we split the dataset into $5+1$ styles --- {\em analogous, complementary, triadic, same, monochromatic} and {\em other}, based on the color theory \cite{malacara2011color}. Table \ref{table:dataset} demonstrates the exact distribution of this dataset.

\noindent\textbf{Polyvore-D.}
Polyvore-D is a subset of the Polyvore dataset where a garment appears in just one split (either ``train'' or ``test''). It is employed to evaluate the generalization ability.

\subsection{Evaluation Metrics}

\begin{figure*}[t]
\begin{center}
   \includegraphics[width=1.0\linewidth]{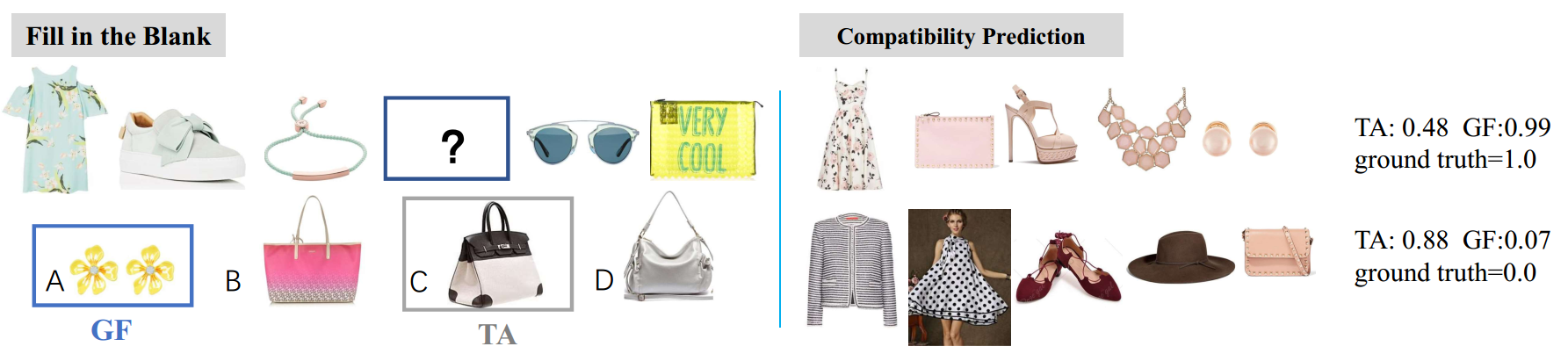}
\end{center}
\vspace{-5pt}
   \caption{\textbf{Visual results on the two main tasks}. Fill-in-the-Blank (\emph{Left}):  the item with the highest compatibility score will be selected as the correct answer. The item in blue box is given by our Graph Filtering (GF) while that in grey box is by Type-Aware (TA)~\cite{VasilevaECCV18FasionCompatibility}. Fashion Compatibility Prediction (\emph{Right}): The scores evaluate the compatibility of the set, higher score means more compatible. Our GF gives more accurate scores than TA.}
\label{fig:evaluation}
\vspace{-2pt}
\end{figure*}

\noindent
\textbf{Fill in the Blank.} It gives a sequence of fashion items, and requires the recommendation system to select one item from four choices that is most compatible with those given items, shown as the left of Figure~\ref{fig:evaluation}. Note that the length of given items is not fixed, ranging from 3 to 16 for selected datasets. The performance is evaluated as the accuracy of correctly answered questions.

\noindent
\textbf{Fashion Compatibility Prediction.} This task requires the collocation system to score a candidate outfit so as to determine if they are compatible and trendy, shown as the right side of Figure~\ref{fig:evaluation}. Performance is evaluated using the area under a receiver operating curve (AUC) \cite{han2017learning}.

\subsection{Comparison Methods}
We compare with Bi-LSTM~\cite{han2017learning}, CSN~\cite{Veit2017Conditional}, TransNFCM~\cite{YangAAAI2019Trans}, Wardrobe~\cite{Hsiao2018Wardrobe} and Type-Aware embeddings~\cite{VasilevaECCV18FasionCompatibility} to verify the superior performance of our graph filtering method on the two tasks.

We also add visual comparison with personalized fashion recommendation methods, \emph{i.e.}, Wardrobe~\cite{Hsiao2018Wardrobe} and Binary-Code~\cite{lu2019learning} to further demonstrate the perceptual appeal of our diversified fashion collocation results via a user study. 
Although they incorporate additional information such as users' shopping history, our recommendations are more favored in user study.

\subsection{Implementation Details}
\noindent
\textbf{Compatibility Learning.} Here only compatibility predictor module is used. For training, our method treats each compatible garment set as a positive example (labelled as 1) and each incompatible garment set as negative (labelled as 0), thus the cross-entropy loss can be used. For testing, each set is scored between 0 and 1 indicating their compatibility.

\noindent
\textbf{Style Learning.}
Here both compatibility predictor and style classifier are used. 
For training, predictor is trained in the same way as stated above, while parameters of style classifier are updated only for compatible garment sets, those labeled as 1, within each batch.
For testing, we only classify the style of an input garment set if its predicted compatibility score is greater than a predefined threshold (\emph{e.g.} $0.5$).

\begin{table*}
\scriptsize
\caption{\emph{Left}: \MakeLowercase{Comparison of Type-Aware (TA)}~\cite{VasilevaECCV18FasionCompatibility} and our Graph Filtering (GF) on Polyvore with different set lengths. GF gains better performance in most cases as the set length increases, that shows its effectiveness in fashion flexibility learning. \emph{Right}: Comparison of Type-Aware (TA)~\cite{VasilevaECCV18FasionCompatibility} and our Graph Filtering (GF) (\emph{with only compatibility predictor used}) on Polyvore and Amazon Fashion with different styles. Our GF gains better performance in most cases, thus it is more suitable for fashion diverse learning.}
\vspace{5pt}
\parbox{0.3\linewidth}{
\centering
\includegraphics[width=1.0\linewidth]{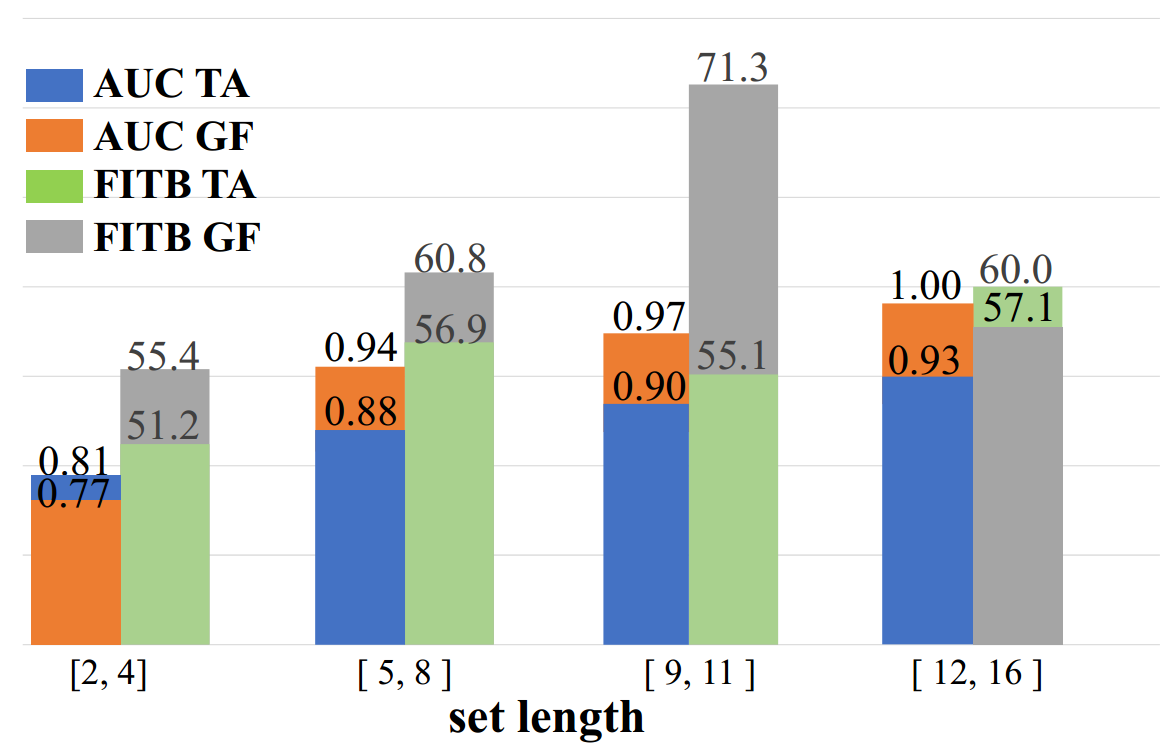}
}
\hfill
\parbox{0.7\linewidth}{
\centering
  \begin{tabular}{c | c c | c c| c c | c c | c c| c c}
    \toprule
    \multicolumn{1}{c}{dataset} &
      \multicolumn{4}{c}{Polyvore} &
      \multicolumn{4}{c}{Polyvore-D} &
      \multicolumn{4}{c}{Amazon}\\
      \midrule
      \multicolumn{1}{c}{} & 
      \multicolumn{2}{c}{AUC} &
      \multicolumn{2}{c}{FITB} &
      \multicolumn{2}{c}{AUC} &
      \multicolumn{2}{c}{FITB} &
      \multicolumn{2}{c}{AUC} &
      \multicolumn{2}{c}{FITB}\\
    \midrule
   {Style} & {TA} & {GF} & {TA} & {GF}  & {TA} & {GF} & {TA} & {GF} & {TA} & {GF} & {TA} & {GF}\\
    \midrule
    Analogous     & 0.86  & \textbf{0.88} & 53.7  & \textbf{56.3} & 0.84 & \textbf{0.91} & 59.1 & 58.4 & 0.98 & 0.98 & 38.3 & \textbf{51.8} \\
    Complementary & 0.86 & \textbf{0.88}  & 55.1  & \textbf{55.4} & 0.85 & \textbf{0.90} & 60.0 & \textbf{60.0} & 0.97 & \textbf{0.98} & 37.5 & \textbf{50.5} \\
    Triadic       & \textbf{0.92} & 0.89 & 56.5 & \textbf{60.8} & 0.86 & \textbf{0.92} & \textbf{64.3} & 57.1 & 0.88 & \textbf{0.97} & 35.9 & \textbf{56.6}\\
    Same          & 0.84 & \textbf{0.87} & \textbf{64.1} &  59.4 & 0.83 & \textbf{0.93} & 60.0 & \textbf{70.3} & 0.95 & \textbf{0.96} & 32.4 & \textbf{53.3}\\
    Monochromatic & 0.83 & \textbf{0.86} & 54.1 & \textbf{53.9} & 0.83 & \textbf{0.85} & 56.5 & \textbf{67.6} & 0.96 & 0.97 & 32.9 & \textbf{52.3} \\
    Other         & \textbf{0.87} & \textbf{0.87}  & 57.1 & \textbf{58.0} & 0.83 & \textbf{0.92} & 60.6 & \textbf{61.9}  & 0.97 & \textbf{0.98} & 34.8 & \textbf{51.7}\\
    \midrule
    Weigthed Avg. & 0.86 & \textbf{0.88} & 55.1 & \textbf{56.9} & 0.84 & \textbf{0.91} & \textbf{59.7} & \textbf{59.7} & 0.95 & \textbf{0.97} & 35.3 & \textbf{52.7}\\
    \bottomrule
  \end{tabular}
}
\vspace{-2pt}
\label{table:comparison} 
\end{table*}

\subsection{Ablation Study}
Here we inspect the individual contribution of each component to the overall performance as shown in Table \ref{table: ablation study}.

\noindent
\textbf{Euclidean Distance.}
It computes the euclidean distance between two garments' embeddings as the pairwise relation.

\noindent
\textbf{Imbalanced Collocation Handling.}
The oversampling method is a straightforward approach to solving the data imbalance for different fashion styles. We use it to balance the number of garment sets per batch, and compare its performance on the two tasks to verify the advantages of our proposed graph filtering on imbalanced learning.
 
\noindent
\textbf{Baseline (Node).}
This baseline processes the feature embedding of each garment (node in the graph) independently without forming garment pair features, and aggregate their features using pooling, followed by MLP for prediction. 

\noindent
\textbf{Baseline (Edge Max/Avg Pooling).}
Here we employ max and average pooling as ``Naive Aggregation'' and compare them with hierarchical feature aggregation to demonstrate the advantage of our proposed graph edge filtering process. Note that our hierarchical feature aggregation does not combine any other pooling strategy.

\begin{figure*}[t]
\centering
\includegraphics[width=1.0\linewidth]{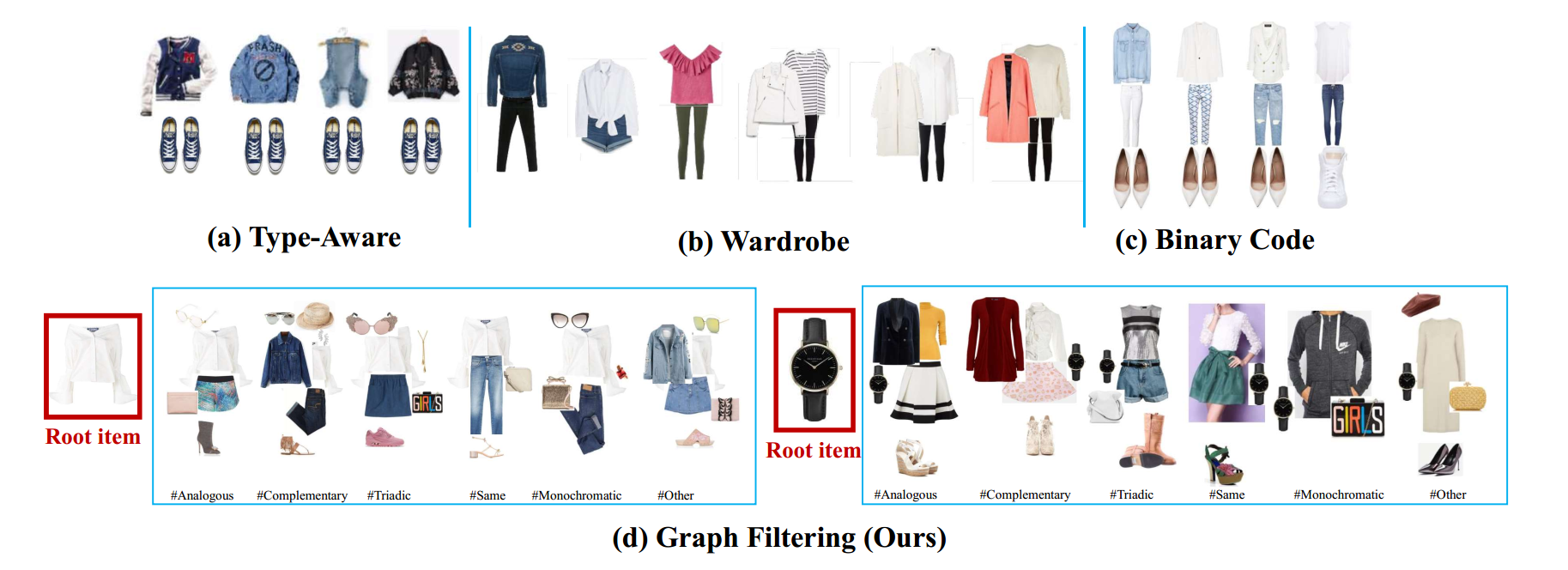}
\caption{ The visual comparison of four state-of-the-art fashion collocation systems. (a) Type-Aware ~\cite{VasilevaECCV18FasionCompatibility} and (c) Binary Code~\cite{lu2019learning} simply recommend the visually similar garments; While (b) Wardrobe~\cite{Hsiao2018Wardrobe} adds user preference information, it still lacks diversity that just recommends two or three garments per set. Our neural graph filtering (d) not only recommends garments together with bags and accessories, but also shows diverse styled collocations.}
\label{fig:diverse}
\end{figure*}

\subsection{Benchmarking Results}

\noindent
\textbf{Polyvore \& Polyvore-D.}
Table \ref{table: ablation study} verifies that our graph filtering method obtains the best performance in both Polyvore~\cite{VasilevaECCV18FasionCompatibility} and Polyvore-D. Statistically, the graph filtering increases compatibility AUC by 8\% and fill-in-blank accuracy by 3.4\% on Polyvore.

We also process each embedding independently and aggregate their features using averaging, shown as ``Embedding Baseline''. It gains 7\% increase on compatibility AUC and 1.2\% increase on fill-in-blank accuracy, demonstrating that modeling pairwise relationship as nonlinear edge feature is more effective than euclidean distance in prior work. Moreover, the comparison of the naive aggregation, ``Max Pooling'' and ``Avg Pooling'', against our graph filtering (hierarchical aggregation) demonstrates that though simply aggregating the edge feature can improve the performance, hierarchically aggregation achieves further improvements.

Table \ref{table:comparison} (\emph{right}) demonstrates that Neural Graph Filtering obtains outstanding performance for various fashion styles. 
Note that most styles obtain evident improvement compared to the Type-Aware method. 

Table \ref{table:comparison} (\emph{left}) shows that Graph Filtering gains better performance in both fill in the blank and compatibility prediction tasks compared to Type-Aware for set lengths.
Note that our model achieves the largest performance gain when the set length ranges between 5 and 11, which are the normal fashion set lengths in daily life. 

\noindent
\textbf{Amazon Fashion.}
Table \ref{table:comparison} (\emph{right most}) compares the results of Type-Aware and Graph Filtering on the dataset Amazon Fashion. Our method achieves 2\% increase on fashion compatibility prediction and 17.4\% increase on fill-in-the-blank. The better performance on Polyvore and Amazon Fashion verifies the effectiveness of our Graph Filtering method.

For the dataset difference, Polyvore and Polyver-D both have more semantic annotations available, e.g., text description, while Amazon Fashion does not. Thus, Type-Aware (TA) shows more disadvantage compared with our Graph Filtering (GF) without the expensive manually constructed textual information(Polyvore). This gap could also validate that our proposed GF framework has better generalization capability. Furthermore, considering most work in fashion compatibility learning just deployed Polyvore, our comprehensive evaluation on the newly assembled “Amazon Fashion Dataset” could alleviate such bias.

\subsection{Diverse Fashion Style Collocation}
Given one fashion item (\emph{e.g.} a blouse) as the ``root'', our fashion recommendation system can generate garment sets of various lengths for a desired style.
It works by iteratively including an item from a candidate pool of each type (\emph{e.g.} tops, bottoms, shoes) till some termination conditions are met.
As Algorithm \ref{code:selection} explains, we first select all the candidates that have compatibility scores greater than a predetermined threshold ($0.5$ in our case) and meanwhile are
classified as the desired style, and then take the item with the highest comparability score.
The process stops if either no candidate exists or the compatibility score including the chosen candidate starts dropping.

\begin{algorithm}[h]
\caption{Item Selection Algorithm}
\label{code:selection}
\begin{algorithmic}[1]
\REQUIRE ~~\\ 
The query item, $x_0$;\\
The number of styles, $N_s$;\\
The number of types, $N_t$;\\
Item subsets, ($X_{{s_0},{t_0}}$,..., $X_{{N_s},{N_t}}$);
\ENSURE ~~\\ 
A set of selected items, $SelectedSets$;
\STATE Initialization: $SelectedSets$=[]
\FOR{each $i \in [1,N_s]$}
\STATE $SetScore = 0$, $Set_{s_i}=[]$;
\FOR{each $j \in [1,N_t]$}
\STATE Select the item $x$ in the subset $X_{{s_i},{t_j}}$ that produces the highest compatibility score, ${ItemScore}$;
\ENDFOR
\IF{$ItemScore$ < $SetScore$}
\STATE Continue the selection for the next type;
\ELSE
\STATE $SetScore$ = $ItemScore$; 
\STATE Add $x$ to $Set_{s_i}$;
\ENDIF
\STATE Add $Set_{s_i}$ to $SelectedSets$;
\ENDFOR
\end{algorithmic}
\end{algorithm}

\begin{table*}
\caption{ResNet18: We use ResNet18 structure as a basis for visual embeddings extraction. Each input is a 224*224*3 RGB image, and the corresponding output is a 512D feature vector. }
    \begin{center}
    \begin{tabular}{c c c c c c c c c}
    \toprule
       \multicolumn{1}{c}{Layer Name} &
       \multicolumn{1}{c}{Conv1} &
       \multicolumn{1}{c}{bn+relu} &
       \multicolumn{1}{c}{Max Pooling} &
       \multicolumn{1}{c}{Conv2} &
       \multicolumn{1}{c}{Conv3*2} &
       \multicolumn{1}{c}{Conv4*2} &
       \multicolumn{1}{c}{Average Pooling} &
       \multicolumn{1}{c}{FC embeds}\\
     \midrule  
      Setting   &  7*7,64 & - & 3*3 & 3*3,64 & 3*3,128 & 3*3,256 & -& -\\
     \midrule
      Output Size   &  56*56 & 56*56 & 28*28 & 28*28 & 14*14 & 7*7 & 1*1 & 512D\\
    \bottomrule
    \end{tabular}
    \end{center}
    \label{tab:resnet18}
\end{table*}

\begin{table*}
\caption{Neural Graph Filtering: The input to the graph network is $n*512$, where $n$ is the number of garment nodes of a set. Within each \textit{Edge Feature Aggregation} operation, $h()$ is implemented by two consecutive $1D \; Conv$, followed by the concatenation, denoted as $\bigoplus$, of a series of pooling operations, including min-, max- and mean-pooling; $g()$ is implemented by two $2D \; Conv$, followed by the concatenation of min- and max-pooling. The \textit{Edge Feature Aggregation} result is then processed by another $1D Conv$, pooling operation, and a series of fully-connected layers to obtain the output logits. }
\scriptsize
    \begin{center}
    \begin{tabular}{c c c c c c}
    \toprule
       {} & 
       Edge Feature Aggregation &
       Edge Feature Aggregation &
       Conv1D &
       Pooling &
       Fully-Connected Layers \\
     \midrule
      \multirow{ 2}{*}{$h()$}  &  1*128, 1*128, & 1*256, 1*256, & \multirow{4}{*}{1*1024} & \multirow{4}{*}{MaxPool $\bigoplus$ MeanPool} & \multirow{4}{*}{(512, 256, 7)} \\ 
      {} & MinPool $\bigoplus$ MaxPool $\bigoplus$ MeanPool & MinPool $\bigoplus$ MaxPool $\bigoplus$ MeanPool  & {} & {} & {} \\
     \cmidrule{1-3}
      \multirow{ 2}{*}{$g()$}  &  1*1*128, 1*1*128, & 1*1*256, 1*1*256, & {} & {} & {} \\
      {} & MinPool $\bigoplus$ MaxPool & MinPool $\bigoplus$ MaxPool & {} & {} & {}\\
     \midrule
     Output Size & n*256 & n*512 & n*1024 & 2048  & 7\\
    \bottomrule
    \end{tabular}
    \end{center}
    \label{tab:dgcnn}
\end{table*}

\subsection{Network Architectures}
We build the visual embedding extraction network on the basis of ResNet18, as shown in Table \ref{tab:resnet18}. Each input is the 224*224*3 RGB image of a garment, and the corresponding output is a 512D feature vector. A single gpu is used to train this convolutional network, and the batch size is set as 240. 
During training, we apply the ``Margin Ranking Loss'', which is mostly used for similarity evaluation. The margin for the triplet loss is set as 0.3. For this visual embedding extraction network, we use ``Adam'' optimization method, and set the learning rate as $5e-5$.

For the proposed neural graph network, to ease the computation, we represent the input nodes of a graph using a 2D matrix with each row vector corresponding to the visual embedding of a node. 
Since the number of nodes across garment sets is unfixed, the number of rows of is then unified to the maximum number of nodes among all the garment sets with proper padding rows filled with zeros. 
We elaborate implementation details of the network structure in Table \ref{tab:dgcnn}.

\section{Human Evaluation.}
Considering fashion is a subjective topic, and it is hard to determine whether a generated fashion set is favorable. Thus, we conduct a user study to investigate the popularity of prior fashion recommendation systems and our {\em Nerual Graph Filtering}.

We randomly surveyed 40 people of different ages \ref{tab:user_age}, sexes \ref{tab:user_gender}, and jobs \ref{tab:user_job}, about 82.5\% surveyees prefer the diverse recommendation results provided by our {\em Nerual Graph Filtering}, shown in Table \ref{table: ablation study} (the column denoted as ``H.''). More than one surveyee point out that such diverse fashion collocation, as Figure \ref{fig:diverse} shows, can concurrently meet their practical needs (to find the suitable garment) and enlighten their fashion minds (to offer diverse styles). Therefore, we can conclude that this diverse fashion collocation framework is well accepted by consumers \wrt the perception of both compatibility and diversity on the recommended fashion sets.

\begin{table}[h]
\caption{User Gender. We randomly surveyed 19 female and 21 male users so as to obtain a balanced gender distribution for our user study.}
\footnotesize
    \begin{center}
    \begin{tabular}{c c c}
    \toprule
    \multicolumn{1}{c}{Gender} &
    \multicolumn{1}{c}{Female} &
    \multicolumn{1}{c}{Male} \\
    \midrule
    number   & 19 & 21 \\
    perfer GF & 17 & 16\\
    \bottomrule
    \end{tabular}
    \end{center}
    \vspace{-10pt}
    \label{tab:user_gender}
\end{table}

\noindent
\textbf{User Gender.}
16 female users and 17 male users \ref{tab:user_gender} express their preferences for the Graph Filtering system rather than others, that shows the popularity and validity of our proposed method.
Female users think these diverse collocations conditioned on one given garment promote the \textbf{functionality} of the garments and also enlighten their \textbf{fashion sense}.
Several male users consider the diversified recommendations will help them break the tedious fashion image and rebuild their \textbf{fashion taste}.

\begin{table}[h]
\caption{User Age. The age distribution of our surveyed users. People between 20 and 40 are usually most enthusiastic to follow the fashion trend, so we focus more on this group.}
    \footnotesize
    \begin{center}
    \begin{tabular}{c c c c c}
    \toprule
    \multicolumn{1}{c}{Age} &
    \multicolumn{1}{c}{$<$20} &
    \multicolumn{1}{c}{20-30} &
    \multicolumn{1}{c}{30-40} &
    \multicolumn{1}{c}{$>$40} \\
    \midrule
     number   & 2 & 19 & 15 & 4  \\
     perfer GF & 1 &17 & 13& 2\\
    \bottomrule
    \end{tabular}
    \end{center}
    \label{tab:user_age}
\end{table}

\noindent
\textbf{User Age.}
Table \ref{tab:user_age} demonstrates the age distribution of our surveyees. We mainly focus on the group aging from 20-year-old to 40-year-old, considering they are most enthusiastic to pursue the fashion trend and also the most potential consumers for online shopping.
Only 4 people in this group did not select ours. While others describe the Neural Graph Filtering recommendation is a \textbf{motivating} system because these diverse collocations highly stimulate their shopping intentions.

\begin{table}[h]
\caption{User Occupations. Our randomly surveyed 40 users have totally different positions in the society.}
\footnotesize
    \begin{center}
    \begin{tabular}{c c c c c c}
    \toprule
    \multicolumn{1}{c}{Jobs} &
    \multicolumn{1}{c}{students}&
    \multicolumn{1}{c}{teachers}&
    \multicolumn{1}{c}{engineers}&
    \multicolumn{1}{c}{staffs}&
    \multicolumn{1}{c}{housewives} \\
    \midrule
     number  & 20 & 5 & 7 & 6 &2  \\
    \bottomrule
    \end{tabular}
    \end{center}
    \label{tab:user_job}
\end{table}

\noindent
\textbf{User Occupation.}
We aim to investigate people of different occupations so as to propose a generally representative user study, explained as Table \ref{tab:user_job}. 
The most frequent reason for the students' favor is that the diversified fashion recommendation help them express their \textbf{characteristics}; the engineers prefer our recommendations because of \textbf{technically} difficult and different visual performance; the staffs recognize these collocations as visually \textbf{professional} and \textbf{effective}.
Therefore, all these distinct reasons considerably verify the validity and popularity of our work.

\section{Conclusion}
In this paper, we propose the \emph{neural graph filtering} framework to enable the flexible and diverse fashion collocation for the first time. 
It not only accepts both inputs/outputs with flexible lengths, but also recommends diverse styled fashion collocations.
Our graph filtering framework exhibits substantial improvements in fashion studies by successfully handling two levels of imbalance in the datasets. It significantly outperforms other state-of-the-art methods on the established tasks ``fashion compatibility prediction'' and ``fill-in-the-blank''.
We also reconstruct a dataset named ``Amazon Fashion'' and divide it into different styles to suit diversity evaluation.
Approximately 82.5\% users prefer our recommendation results, demonstrating such diverse fashion collocation meets their realistic requirements as well as offers them fashion enlightenment.

\appendix

\begin{figure*}[t]
\begin{center}
 \includegraphics[width=1\linewidth]{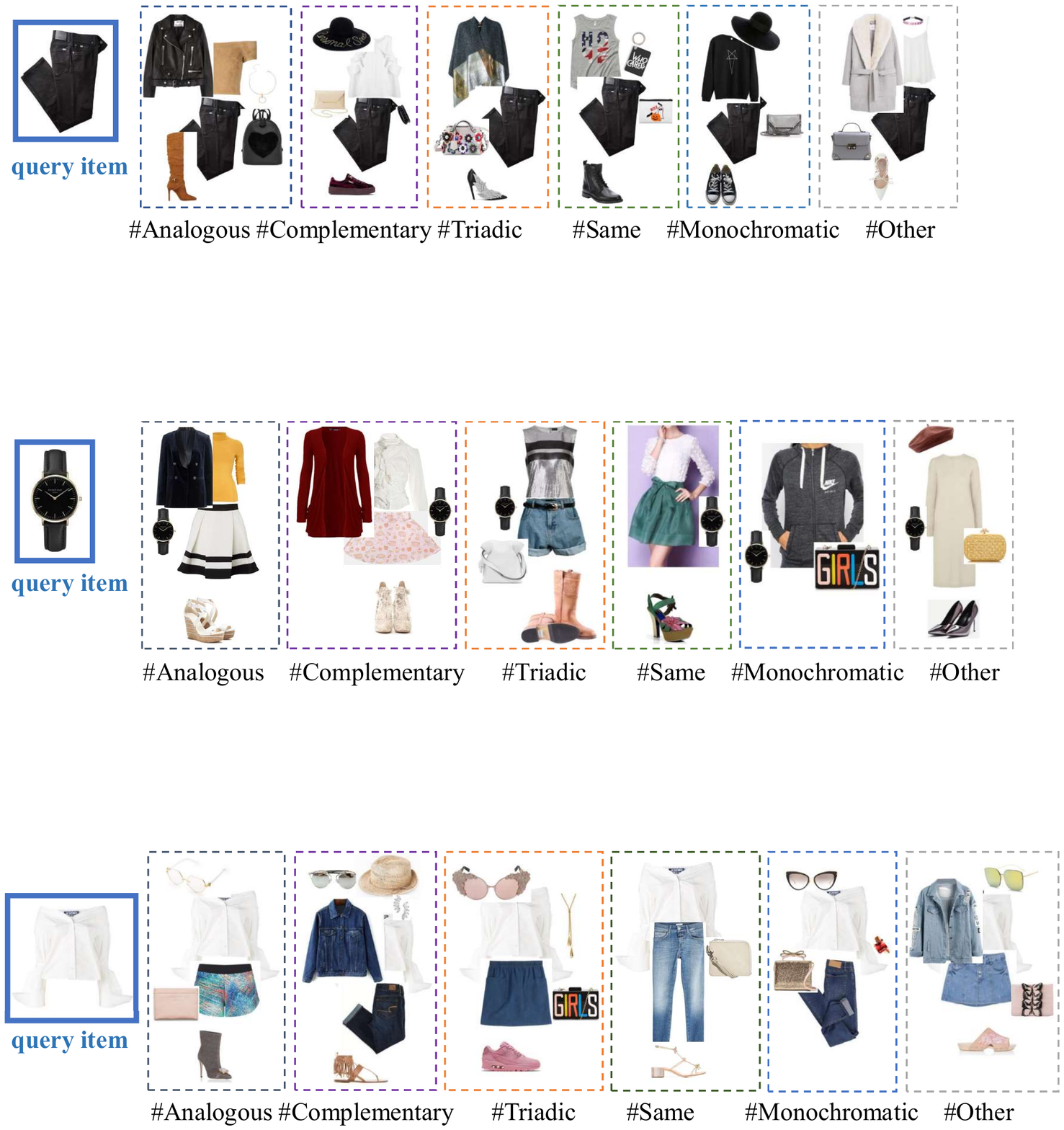}
\end{center}
\vspace{-30pt}
\caption{Visual results of diverse fashion collocation on Polyvore dataset (part 1).}
\label{fig:Polyvore_Gen1}
\vspace{-5pt}
\end{figure*}

\begin{figure*}[t]
\begin{center}
 \includegraphics[width=1\linewidth]{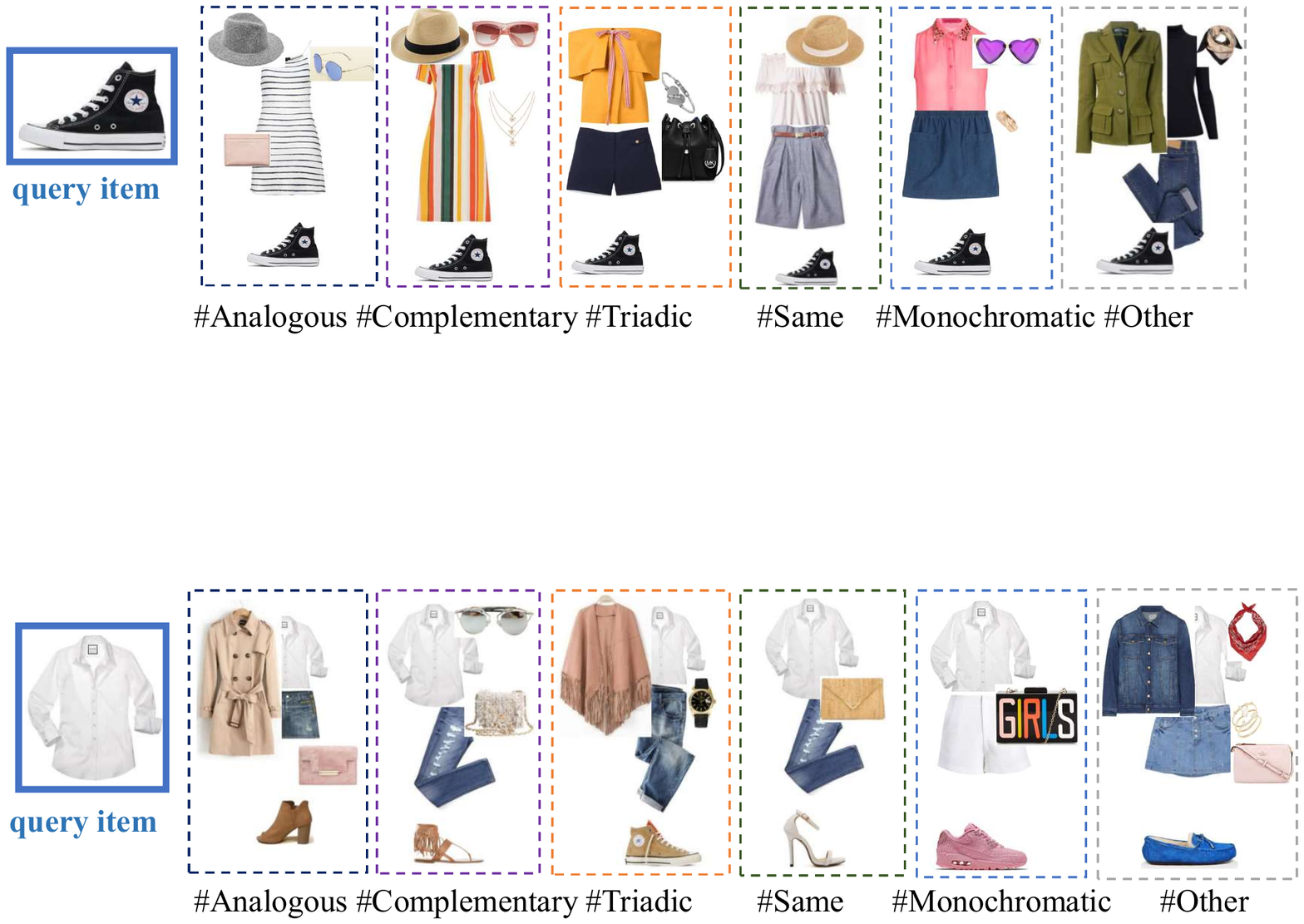}
\end{center}
\vspace{-30pt}
\caption{Visual results of diverse fashion collocation on Polyvore dataset (part 2).}
\label{fig:Polyvore_Gen2}
\vspace{-5pt}
\end{figure*}

\begin{figure*}[t]
\begin{center}
 \includegraphics[width=1\linewidth]{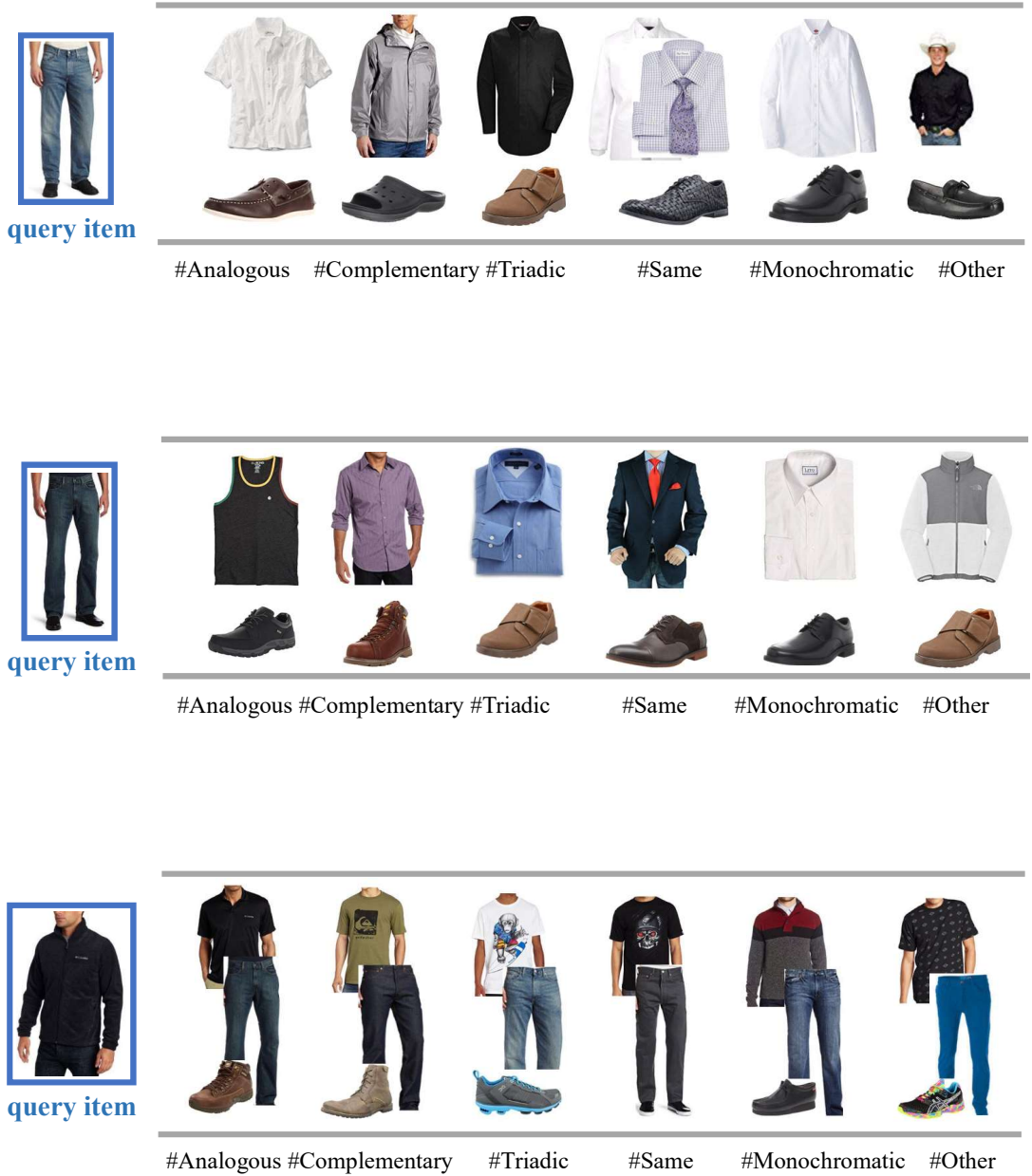}
\end{center}
\vspace{-30pt}
\caption{Visual results of diverse fashion collocation on Amazon Fashion dataset (part 1).}
\label{fig:Amazon_Gen1}
\vspace{-5pt}
\end{figure*}

\begin{figure*}[t]
\begin{center}
 \includegraphics[width=1\linewidth]{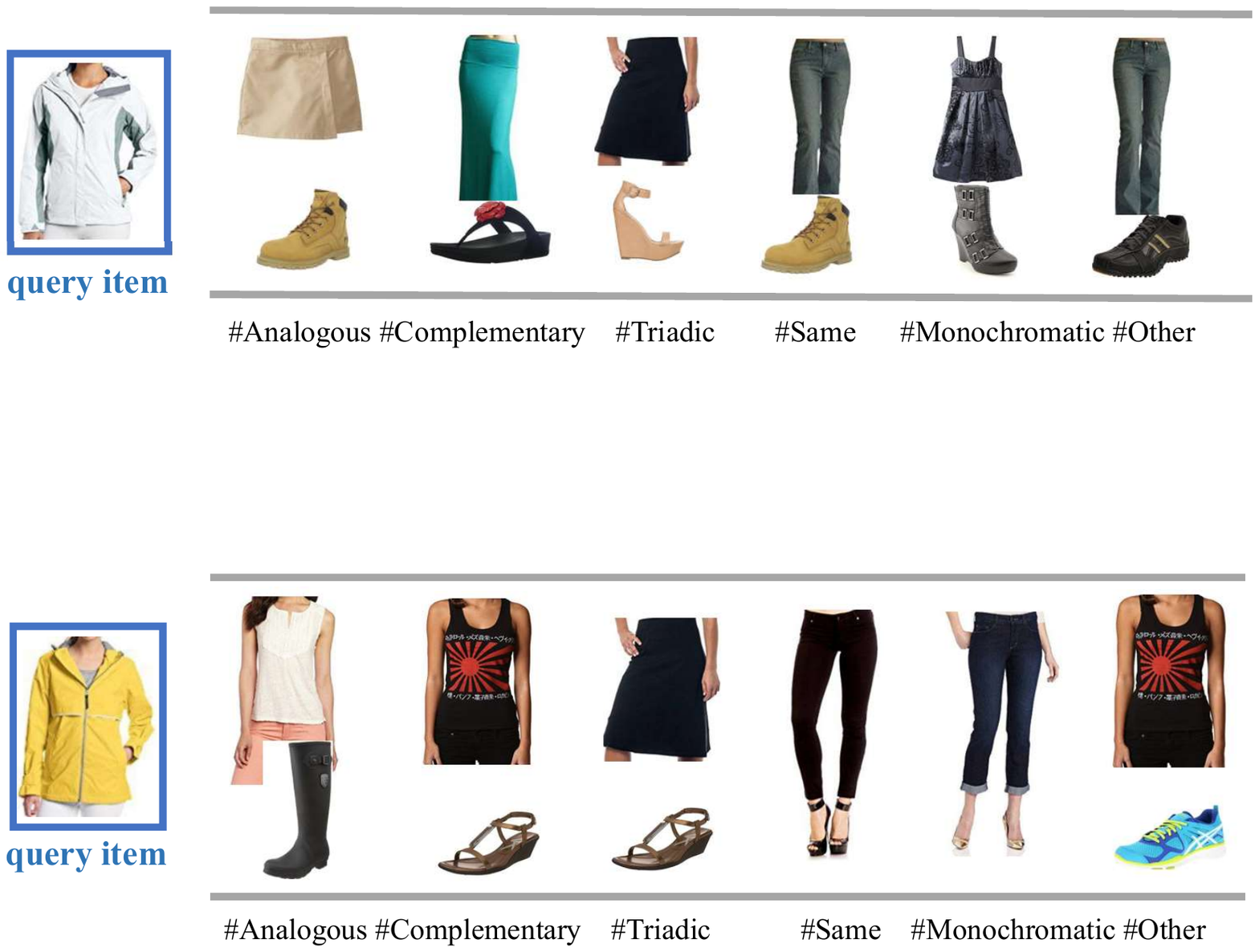}
\end{center}
\vspace{-30pt}
\caption{Visual results of diverse fashion collocation on Amazon Fashion dataset (part 2).}
\label{fig:Amazon_Gen2}
\vspace{-5pt}
\end{figure*}

The visual results of diverse fashion collocation are shown in Figure \ref{fig:Polyvore_Gen1}, Figure \ref{fig:Polyvore_Gen2}, Figure \ref{fig:Amazon_Gen1} and Figure \ref{fig:Amazon_Gen2}. 
Either given the common fashion item like a shirt, a jeans or the shoes, or the difficult accessory item like the watch, our Neural Graph Filtering method can produce diverse and compatible fashion sets.

\ifCLASSOPTIONcaptionsoff
  \newpage
\fi

{
\bibliographystyle{IEEEtran}

}

\end{document}